# Automatic and Accurate Classification of Hotel Bathrooms from Images with Deep Learning

**Hakan Temiz[1]***

[1]*Artvin Çoruh University, Department of Computer Engineering, 08000, Artvin, TURKİYE*

**Abstract**

Hotel bathrooms are one of the most important places in terms of customer satisfaction, and where the most complaints are reported. To share their experiences, guests rate hotels, comment, and share images of their positive or negative ratings. An important part of the room images shared by guests is related to bathrooms. Guests tend to prove their satisfaction or dissatisfaction with the bathrooms with images in their comments. These Positive or negative comments and visuals potentially affect the prospective guests. In this study, two different versions of a deep learning algorithm were designed to classify hotel bathrooms as satisfactory (good) or unsatisfactory (bad, when any defects such as dirtiness, deficiencies, malfunctions were present) by analyzing images. The best-performer between the two models was determined as a result of a series of extensive experimental studies. The models were trained for each of 144 combinations of 5 hyper-parameter sets with a data set containing more than 11 thousand bathroom images, specially created for this study. The "HotelBath" data set was shared also with the community with this study. Four different image sizes were taken into consideration: 128, 256, 512 and 1024 pixels in both directions. The classification performances of the models were measured with several metrics. Both algorithms showed very attractive performances even with many combinations of hyper-parameters. They can classify bathroom images with very high accuracy. Suh that the top algorithm achieved an accuracy of 92.4% and an AUC (area under the curve) score of 0.967. In addition, other metrics also proved the success of the algorithm. The proposed method can allow the rapid, accurate and automatic detection of such undesired circumstances in hotel bathrooms from images. Such a detection system can allow hotel management to take necessary actions quickly to remedy such unsatisfactory cases.

*\*Corresponding Author: hakan_temiz@hotmail.com*

## 1. Introduction

While increasing the number of loyal customers, guest satisfaction can play an important role in reaching potential customers with recommendations. It is a known fact that the complaints of the dissatisfied customer are more effective on the potential customers than the recommendations of satisfied customers (Genç and Batman, 2018). Many factors determine the satisfaction of the guests. The main ones are environment, facilities, staff and services and price restaurants, food and beverages and security, staff communication, hotel image, price, transportation services and restaurants (Liu et al., 2013). Customer complaints are a fact of life in the hospitality industry. Because from time to time mistakes are almost inevitable. Complaints include a wide variety of complaints, often covering many issues such as service delivery, equipment failure, and personnel failure (Levy et. al., 2012).

Hotel guest satisfaction is greatly affected by the guest room experience. Because the guest room is the main attractive and central feature of a hotel (Cheung, 2002). The physical aspects and tangible dimensions of the guest room are the keys to guest satisfaction and the most important consideration for return patronage (Ogle, 2009). Several studies have found that guest rooms make a more lasting impression on the guests than the exterior architecture, lobby or any other interior space (DeVeau, 1996; Penner et al., 2013). Prasad, Wirtz, and Yu found that 55% of problems in hotels were service- or facility-related problems, and, 12% of these problems were related to the bathroom (Prasad et. al., 2014).

Guests share their satisfaction and dissatisfaction on platforms such as hotel review pages, tourism content providers, and rating sites (TripAdvisor, Trivago, etc.). Sometimes they just score, sometimes add comments and even add visuals of their experiences. To maintain their existence and provide a competitive advantage, hotel businesses need to be able to respond quickly and take quick action to the ideas, comments and complaints of the guests, in addition to producing goods and services for changing needs and requests to ensure maximum customer satisfaction. In this context, there is a need to design systems that can detect potentially unsatisfactory experiments or problems from online content and enable remedying actions to be taken. Although a very simple solution is the textual analysis of guest comments, in this study, unlike traditional approaches, a system has been developed that enables the problems related to hotel bathrooms to be detected from the images. According to the research, an autonomous problem detection system similar to the one in this study has not been encountered.

This study aimed to develop a method that makes an automatic classification of the condition of the bathroom from the bathroom images shared by the guests, as problematic (bad) or problem-free (good). As a result of the studies, a deep learning model, which was originally designed for super resolution problems, was modified to meet the classification problem and expected to predict the given bathroom images in two classes: good or bad. For this purpose, further modifications were made to improve the performance of the model while lightening the architecture. The results revealed that the designed model can automatically detect situations that may cause problematic or negative interpretations from the images quite successfully. In addition, within this study, a novel dataset, "HotelBath" (Temiz, 2022), was created and introduced in the experiments and shared with the research community.

## 2. Literature

Many important criteria that can be called a problem in the guest room. Bathroom, size, fragrance, bed, television, decor, air conditioning, carpet and floor, minibar and refrigerator, furniture, windows, lighting, kitchen, etc. examples of these can be given. Among these, one of the most complained about is the bathroom. The study by Levy, Duan and Boo (2013) revealed that almost one-fifth of customer complaints are related to the bathroom, depending on the type and class of the hotel. Typical bathroom complaints include the size and functionality of the room and issues with the shower, tub, sink, and toilet. Susskind and Verma (2011) tested the effects of changes in bathroom lighting on guest satisfaction, demonstrating how important even lighting is. Another (Prayukvong et al., 2007) study found that negative customer experience, especially in 1-2 star category hotels, was mainly caused by bathroom and toiletries. When another study makes an evaluation in terms of cleanliness among the factors affecting the choice of accommodation, it has been determined that the two most important areas are "Bathroom and Toilet" and "Kitchen" and that these are much more important than the cleanliness of other areas (Lockyer, 2003). Genç and Batman (2018) have stated that the bathroom and toilet are at the top of the issues that tourists complain about the most in historical mansion businesses. Gahramanov and Türkay (2019) found that the bathroom ranked fourth in the tourist opinions they received about the 40 most important factors in hostel management.

In the literature, various studies classify interior parts (room, kitchen, bathroom, toilet, etc.). Yang, Rangarajan and Ranka (2018) used the Global Interpretation via the Recursive Partitioning (GIRP) method, similar to the CART (classification and regression tree) algorithm, which is a machine learning algorithm. Wu, Christensen and Rehg (2009) used the Bayesian filtering approach. On the other hand, Kumari and Maan (2020) proposed a convolutional neural network (CNN) based deep learning model for spatial classification. Swadzba and Wachsmuth (2014) proposed a method introducing the gist feature vector of images with the RANSAC algorithm (RANdom SAmple Consensus). However, all of these studies are only for the determination of space sections separately from each other.

In this study, based on deep learning, a method has been developed that automatically detects whether the bathroom is problematic/problematic or good from the bathroom images uploaded by the guests to a content site. In the literature, similar to this study; Based on the automatic analysis of the image, no study has been found that detects a problematic or problematic or problem-free bathroom.

## 3. Materials and Methods

### 3.1. Dataset

Bathroom images of 966 hotels in the Antalya region (uploaded to the site by the hotel and guests) were downloaded from TripAdvisor.com with a crawler program specifically developed for this project. In total, 11,116 images were collected from the website. The collected bathroom images were manually labeled as good and bad. Broken, cracked, faulty plumbing or equipment, or dirty, messy, unemptied washbasin, peeled off paint, wall or ceiling, etc. images that indicate possible problems were tagged as bad; while clean, tidy, hygienic etc. images pointing to a satisfying bathroom environment were also labeled as good. Some examples of bathroom images that were qualified as good and bad are given in Figure 1. This dataset (Temiz, 2022), entitled "HotelBath", is shared with the research community within this study.

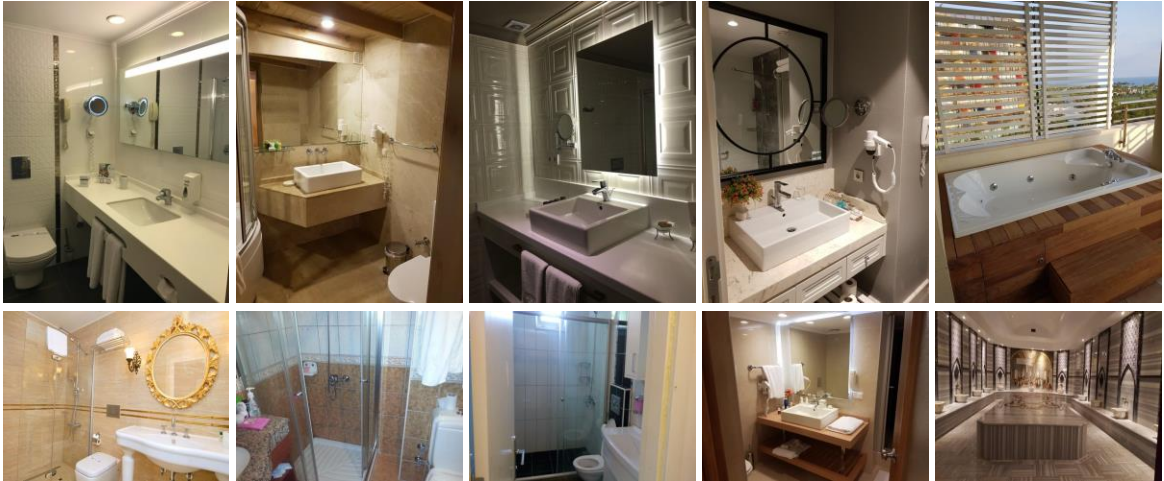

**(a)** Sample bathroom images labeled as good (satisfying)

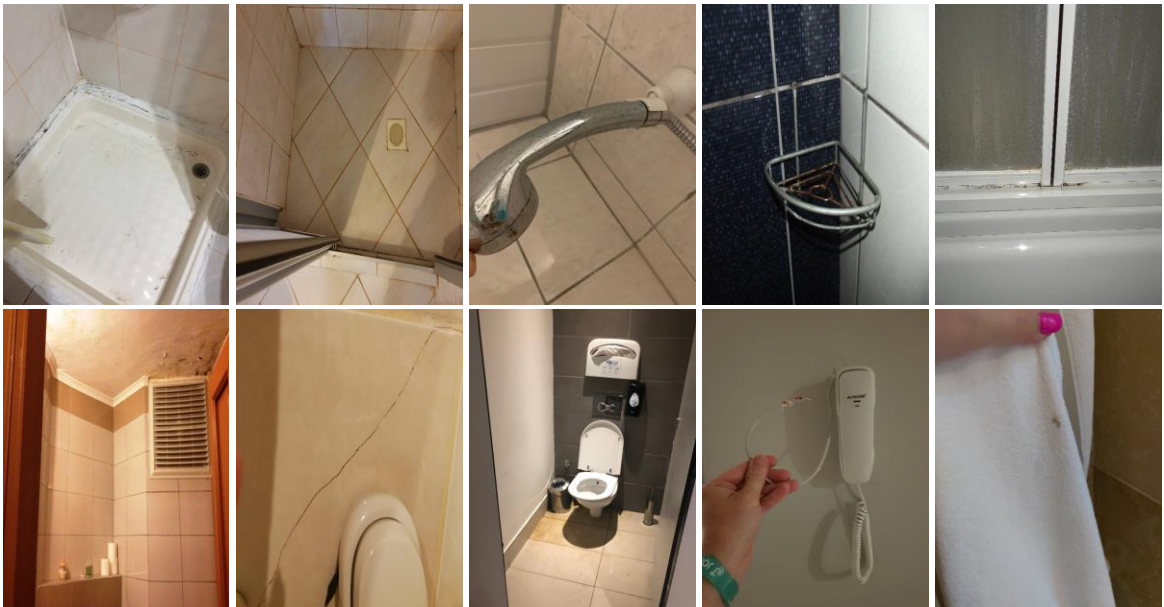

**(b)** Example bathroom images labeled as bad (problematic/unsatisfying)

**Figure 1.** Some example bathroom images considered as good (a) or bad (b).

As a result of labeling, 7181 bathroom images were classified as good and 3935 images as bad. Of the good and bad images, 6822 and 3739 were reserved for the training, and 359 and 196 were reserved for testing, respectively. The ten percent of the training data is reserved for the validation of the models. The details of the entire are given in Table 1.

**Table 1.** Details of the Image Dataset

|  | **Good** | **Bad** | **Total** |
|---|---|---|---|
| Training | 6822 | 3739 | 10561 |
| Test | 359 | 196 | 555 |
| Total | 7181 | 3935 | 11116 |

### 3.2. Model

In this study, the DECUSR model (Temiz and Bilge, 2020), which was originally designed for the super resolution problem, was modified to be used in the classification problem. In this context, a Dense(1) layer with a single neuron and activation function 'sigmoid' has been added to the end of the network. A Flatten() layer is placed in front of the Dense(1) layer to reduce the output to one dimension as the Conv2D(1,1) layer, which was the last layer in the original version, produces a two-dimensional output. Unlike the original version, MaxPooling2D and BatchNormalization layers are optionally placed behind the repeating blocks of the network. The BatchNormalization layer is only applied to the last layer of repeating blocks (RBs) if added to the network. The MaxPooling2D layer is applied to the feature upscaling layer (LFUP) and the last layer of RBs. If one or both of the BathNormalization or MaxPooling2D layers are present, the final layers of the respective RBs do not output directly to the successive concatenation layers. Instead, the outputs of the relevant final layers are first given to the BatchNormalization and/or MaxPooling2D layers, and the outputs are then given to the further concatenation layers.

In addition to the modified DECUSR model, a lighter version, DECUSR-L, was also designed with the hope of achieving better classification performance with a lighter model. The two upsampling layers —LFUP and LDUP (direct upscaling layer)— of the original DECUSR are omitted in the lightweight model as the original DECUSR architecture was designed to enlarge the input image in-house and such a function would not be needed in the classification problem. Thus, a lighter model with fewer parameters was designed. Also, apart from the modified DECUSR model, a MaxPooling2D and BatchNormalization layer is placed at the end of the feature extraction (consisting of the first four layers) block. The remaining structure of the network is the same as in the other model. While the modified DECUSR model has 45,122 parameters, the DECUSR-L has 11,138 parameters. Figure 2 shows the architecture of both models. The optional layers—BatchNormalization and MaxPooling2D— are shown with dashed lines.

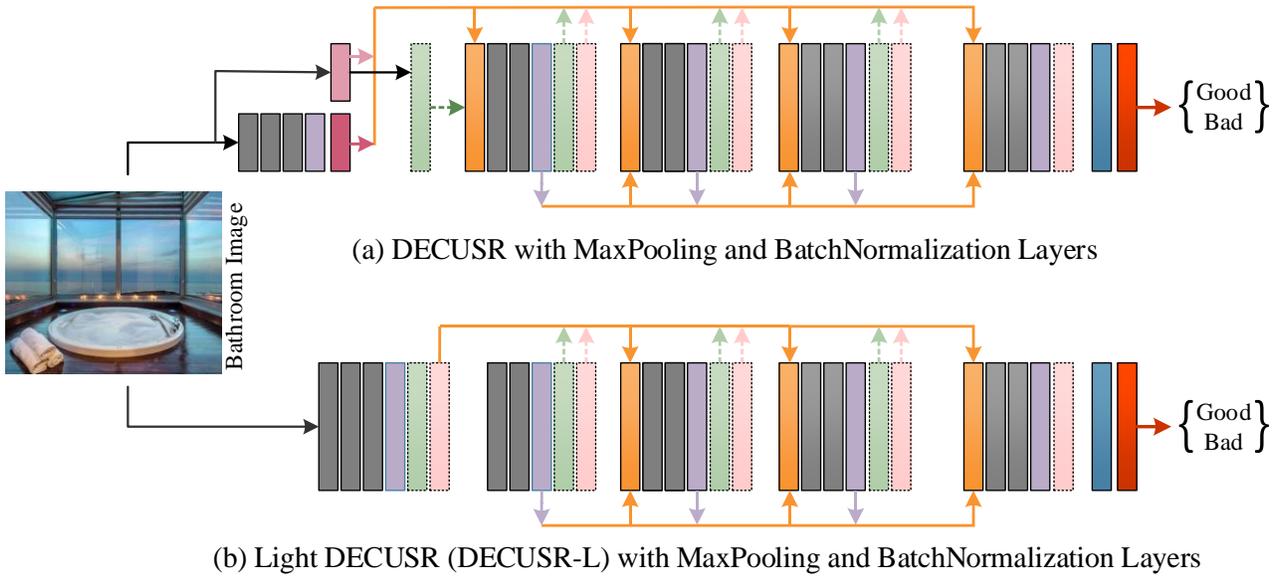

(a) DECUSR with MaxPooling and BatchNormalization Layers

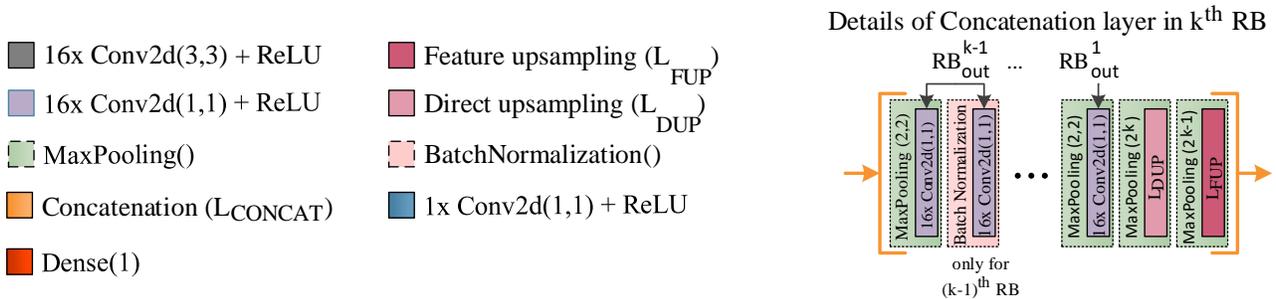

(b) Light DECUSR (DECUSR-L) with MaxPooling and BatchNormalization Layers

**Figure 2.** Details of both models: DECUSR and DECUSR-L. The optional layers, MaxPooling and BatchNormalization, are shown with dashed lines.

### 3.3. Training

A two-phased process was adopted to determine (find) the most successful model with optimal hyper-parameters. In the first phase, both models were trained for 50 epochs at most for each of 144 different combinations of the 5 main hyper-parameters. After the accomplishment of the first phase, the top two of both models with the best classification performance were taken to be trained extensively in the second phase.

In the first phase, the following hyper-parameters were used: Optimizer, Activation, Learning rate, MaxPooling and Batch Normalization. Three optimizers—RMSProp, ADAM and SGD— were used. The Elu, Exponential, Selu, Relu, Sigmoid and Tanh were used as activation functions. The training started with two different versions of the learning rate: 1E-3 and 1E-4. The binary cross-entropy was used as the loss function in the experiments in both phases. The trainings were conducted taking into account the application and non-application of the max pooling and batch normalization processes. The image size was chosen as 256x256 pixels as the reference size in the first phase of the experiment.

The studies conducted in the first phase have shown that the RMSProp and ADAM optimizers are significantly successful for both models. The top four models (two for each of both versions) generally have the learning rate of 1E-3 and the Relu activation function. All of these successful models have MaxPooling layers. In this context, it is seen that MaxPooling directly contributes positively to performance. The same cannot be said for the Batch Normalization layer. Half of the four successful models from the first phase have this layer, while the other half do not. All training as well as test work was done on a computer with NVIDIA GEFORCE RTX 2080 GPU and AMD Ryzen Threadripper Pro Series CPU. All developments and experiments were conducted with Keras with the TensorFlow backend.

The image dimension that gives the best performance in comprehensive training is also included in the analysis. In this context, in addition to the previous image size, the trainings were performed for each of 128x128, 512x512 and 1024x1024 pixel dimensions for 500 epochs at most in the second phase. In this way, the model that gave the final best performance was obtained. Figure 3 shows in detail the entire training workflow. The parameters used in the trainings, the best four models obtained in the first stage and the information on the final best model are given in the figure.

### 3.4. Test

A total of 555 images (359 labeled as good, and 196 as bad) were used to evaluate the classification performance of the top four models. The following measures are exploited to assess the performances: TP, FP, FN, TN, Accuracy, Precision, Recall, and AUC. Their explanations are given below:

**True Positive (TP)**: Values that are actually positive and predicted positive.

**False Positive (FP)**: Values that are actually negative but predicted to be positive.

**False Negative (FN)**: Values that are actually positive but predicted to be negative.

**True Negative (TN)**: Values that are actually negative and predicted to be negative.

**Accuracy**: The ratio of all correctly predicted positives and negatives.

$$\frac{TP + TN}{TP + TN + FP + FN} \tag{1}$$

**Precision**: The ratio of the number of correctly predicted positives to all predicted positives. It is a measure of how many of those predicted positively are actually correct.

$$\frac{TP}{TP + FP} \tag{2}$$

**Recall**: The ratio of the number of correctly predicted positives to the true positives. It is a measure of how many of the true positives are correctly predicted. Also known as specificity.

$$\frac{TP}{TP + FN} \tag{3}$$

**TPR**: True positive rate. It is another synonym for Recall and is defined in the same way as the Recall given in (3).

**FPR**: False positive rate. The ratio of the number of falsely predicted positives to the true negatives. Also called the probability of a false alarm.

$$\frac{FP}{FP + TN} \tag{4}$$

**AUC**: Area under the receiver operating characteristic (ROC) curve. AUC measures the entire two-dimensional ROC curve which is calculated over the ratio of TPR to FPR for all possible classification thresholds. The closer to 1.0, the higher the classification excellence.

| Model | Optimizer | Activation | Learn. Rate | Max Pooling | Batch Norm. |
|---|---|---|---|---|---|
| DECUSR, DECUSR-L | RMSProp, ADAM, SGD | Elu, Exponential, Selu, Relu, Sigmoid, Tanh | 1E-3, 1E-4 | Yes, No | Yes, No |

Train both models for each of the 144 parameter combinations (image size: 256)

First phase
Second phase

The top two parameter combinations of both models | | | | | | Accuracy | Image Size

| | | | | | | Accuracy | |
|---|---|---|---|---|---|---|---|
| DECUSR | ADAM | Relu | 1E-4 | Yes | No | 0.881 | 128 |
| DECUSR | RMSProp | Relu | 1E-3 | Yes | Yes | 0.885 | 256 |
| DECUSR-L | ADAM | Relu | 1E-3 | Yes | Yes | 0.915 | 512 |
| DECUSR-L | RMSProp | Elu | 1E-3 | Yes | No | 0.906 | 1024 |

Exhaustive training of models to determine the best model and image size

| **DECUSR-L** | **RMSProp** | **Elu** | **1E-3** | **Yes** | **No** | **0.924** | **512** |
|---|---|---|---|---|---|---|---|

The best model and parameter set

**Figure 3.** Training phases. Both models were trained for 144 combinations of 5 hyper-parameters. The top two ones of both models were introduced to an exhaustive training phase for different image sizes.

## 4. Results and Discussion

The numerical results obtained in the 2$^{nd}$ stage of the training are given in Table 2. The table presents the four models that were successful in Stage 1 with their parameter sets. In addition, Accuracy, Precision, Recall, AUC, TP, TN, FP, and FN values are also presented in the table. The obtained performances are given by grouping according to the image size.

According to the table, it is seen that the classification performance of all models in general is remarkably successful. When the optimizers are compared to each other, it can be said that RMSProp performs better than ADAM, although there is not a significant performance difference. In terms of the activation functions, the Elu appears to be better than the Relu. The learning rate of 1E-3 offered more successful results than that of 1E-4. It is seen also that applying batch normalization slightly contributes positively to the performance when the activation function Relu is selected, and vice versa. However, it is not possible to say that technique always makes a positive contribution. It is also seen from the table that better results are obtained in some cases where this method is not applied. The Batch Normalization layers normalize the layer weights according to the statistical distribution of the weights. This may cause sometimes the loss of the information grasped and stored by the weights about the distinguishing and characteristic features in the images. However, it is easily seen from the table that the max pooling technique was applied to all models. So, the MaxPooling layers also positively contributed to the classification performance of the models. This may be because these layers contribute to amplifying high-frequency features and distinguishing details rather than fairly smooth and/or low-frequency information in images.

When we examine the table carefully, it becomes clear that DECUSR-L outperforms DECUSR for all image sizes. The lowest classification performance was obtained from the DECUSR+ADAM optimizer pair with an image size of 1024 pixels. DECUSR achieved the highest classification accuracy with RMSProp optimizer: 88.5% for both 256x256 and 512x512 pixel image sizes. However, when the other criteria are taken into account—e.g., AUC is 0.932 for 256x256 and 0.916 for 512x512—it is seen that the image size of 256x256 pixels provides more successful results. With a few minor exceptions, in general, the image size of 512x512 pixels yielded much better results than other image sizes.

As a result, the most successful model, DECUSR-L with RMSProp optimizer, achieved 92.4% classification accuracy with the following parameter sets: optimizer is RMSProp, activation is Elu, learning rate is 1E-3, Max Pooling is applied, Batch Normalization is not applied, and the image size is 512x512. This model is marked with bold text in the table. As can be seen from the table, the most successful model showed very good classification success by reaching the Precision, Recall and AUC values of 0.930, 0.955, and 0.967, respectively. The AUC value of 0.967 is a very good indication of the excellence of the model in the classification. The second-best performance was achieved again by DECUSR-L but with ADAM optimizer. The average prediction time of DECUSR and DECUSR-L are 5.49E-05 and 4.79E-05 seconds. So, it is seen how successful and faster a model with much fewer parameters can be when it is well-designed according to the subject studied.

Not only did the top-performing model perform enticingly, but other models also achieved highly competitive results. The difference in the performances of the top model and the others (especially the second model) is very small. This is a good indication that the model's ability to classify images (or detect the difference) is quite good even with different sets of hyper-parameters. This means that the architecture of the network is well suited to this classification problem, apart from the hyper-parameters.

**Table 2.** Results of the best two parameter sets of both models: DECUSR and DECUSR-L. The bold text belongs to the best model and parameter set.

| Model | Optimizer | Activation | L.Rate | Max Pool. | Batch N. | Loss | Accuracy | Precision | Recall | AUC | TP | TN | FP | FN |
|---|---|---|---|---|---|---|---|---|---|---|---|---|---|---|
| | | | | | Image Size: (128x128) | | | | | | | | | |
| DECUSR | ADAM | Relu | 1E-04 | Yes | No | 0.307 | 0.874 | 0.885 | 0.925 | 0.938 | 332 | 153 | 43 | 27 |
| DECUSR | RMSProp | Relu | 1E-03 | Yes | Yes | 0.305 | 0.872 | 0.881 | 0.928 | 0.941 | 333 | 151 | 45 | 26 |
| DECUSR-L | ADAM | Relu | 1E-03 | Yes | Yes | 0.284 | 0.892 | 0.903 | 0.933 | 0.944 | 335 | 160 | 36 | 24 |
| DECUSR-L | RMSProp | Elu | 1E-03 | Yes | No | 0.254 | 0.910 | 0.919 | 0.944 | 0.958 | 339 | 166 | 30 | 20 |
| | | | | Image Size (reference): (256x256) | | | | | | | | | | |
| DECUSR | ADAM | Relu | 1E-04 | Yes | No | 0.333 | 0.881 | 0.885 | 0.939 | 0.933 | 337 | 152 | 44 | 22 |
| DECUSR | RMSProp | Relu | 1E-03 | Yes | Yes | 0.374 | 0.885 | 0.885 | 0.944 | 0.932 | 339 | 152 | 44 | 20 |
| DECUSR-L | ADAM | Relu | 1E-03 | Yes | Yes | 0.251 | 0.915 | 0.924 | 0.947 | 0.955 | 340 | 168 | 28 | 19 |
| DECUSR-L | RMSProp | Elu | 1E-03 | Yes | No | 0.254 | 0.906 | 0.918 | 0.939 | 0.954 | 337 | 166 | 30 | 22 |
| | | | | | Image Size: (512x512) | | | | | | | | | |
| DECUSR | ADAM | Relu | 1E-04 | Yes | No | 0.408 | 0.832 | 0.887 | 0.850 | 0.900 | 305 | 157 | 39 | 54 |
| DECUSR | RMSProp | Relu | 1E-03 | Yes | Yes | 0.758 | 0.885 | 0.902 | 0.922 | 0.916 | 331 | 160 | 36 | 28 |
| DECUSR-L | ADAM | Relu | 1E-03 | Yes | Yes | 0.225 | 0.921 | 0.936 | 0.942 | 0.964 | 338 | 173 | 23 | 21 |
| **DECUSR-L** | **RMSProp** | **Elu** | **1E-03** | **Yes** | **No** | **0.215** | **0.924** | **0.930** | **0.955** | **0.967** | **343** | **170** | **26** | **16** |
| | | | | | Image Size: (1024x1024) | | | | | | | | | |
| DECUSR | ADAM | Relu | 1E-04 | Yes | No | 0.487 | 0.791 | 0.800 | 0.903 | 0.847 | 324 | 115 | 81 | 35 |
| DECUSR | RMSProp | Relu | 1E-03 | Yes | Yes | 0.497 | 0.843 | 0.876 | 0.883 | 0.908 | 317 | 151 | 45 | 42 |
| DECUSR-L | ADAM | Relu | 1E-03 | Yes | Yes | 0.233 | 0.919 | 0.927 | 0.950 | 0.962 | 341 | 169 | 27 | 18 |
| DECUSR-L | RMSProp | Elu | 1E-03 | Yes | No | 0.240 | 0.915 | 0.917 | 0.955 | 0.957 | 343 | 165 | 31 | 16 |

## 5. Conclusion

In this study, a method is proposed to classify hotel bathrooms as problematic (unsatisfactory or bad) and problem-free (satisfactory or good) based on automatic analysis of the images. For this purpose, two different architecture of a deep learning model whose success has been proven in the literature has been designed to classify the bathroom images. The models were trained and tested for each set of 144 combinations of the 5 most important hyper-parameters to determine the best-performing architecture and hyper-parameter set. Four different image sizes were taken into consideration: 128x128, 256x256, 512x512 and 1024x1024 pixels. A novel dataset, the "HotelBath", was created and used in the training, test and validation processes of the models. It is shared with the research community within this study as well. Several classification metrics were used to assess the classification performances of the models.

The results showed that both architecture offered very attractive classification performance for many hyper-parameter combinations. However, of the two models, the one with fewer parameters performed better than the other. In general, the image size of 512x512 pixels yielded better classification performance than the other sizes. The best-performing model was trained with the following other hyper-parameter values: optimizer is RMSProp, activation is Elu, learning rate is 1E-3, Max Pooling is applied, and Batch Normalization is not applied. The top model was able to detect with a high 92.4% accuracy that a bathroom is either problematic (dissatisfying customers) or uneventful (satisfying customers) from the images. Besides, other metrics also prove the success of the algorithm with astonishing high scores. E.g., the model obtained a 0.967 AUC score.

With the proposed system, it has been shown that the problematic bathrooms that cause or may cause the guest's dissatisfaction (or in other respects, support their satisfaction) can be automatically detected from images. This algorithm can be used to design a system that can automatically and accurately detect undesired defects and problems in hotel bathrooms. Such an automatic detection system will enable hotel managers to quickly and easily detect undesirable, dissatisfied situations or situations that may lead to negative comments from guests, and thus, take remedying actions.

**Acknowledgment**

This study was presented at the 1st International Conference on Scientific and Academic Research ICSAR 2022